\documentclass[runningheads]{llncs}

\usepackage[year=2024,ID=10]{eccv}

\usepackage{comment}
\usepackage{xcolor}
\usepackage{graphicx}
\usepackage{multirow}
\usepackage{soul}
\usepackage{array}
\usepackage{caption}
\newcolumntype{?}{!{\vrule width 2pt}}

\usepackage{eccvabbrv}

\usepackage{graphicx}
\usepackage{booktabs}

\usepackage[accsupp]{axessibility}  

\usepackage{hyperref}

\usepackage{orcidlink}
\usepackage{array, makecell} 

\begin{document}

\title{Generalizability analysis of deep learning predictions of human brain responses to augmented and semantically novel visual stimuli} 

\titlerunning{Generalizability analysis of DL predictions of human brain responses}

\author{Valentyn Piskovskyi\inst{1}\orcidlink{0009-0002-0524-340X} \and
Riccardo Chimisso\inst{1}\orcidlink{0009-0002-0962-1404} \and
Sabrina Patania\inst{1}\orcidlink{0000-0001-7279-723X} \and \\
Tom Foulsham\inst{2}\orcidlink{0000-0002-8444-7269} \and 
Giuseppe Vizzari\inst{1}\orcidlink{0000-0002-7916-6438} \and 
Dimitri Ognibene\inst{1, 2}\orcidlink{0000-0002-9454-680X}}

\authorrunning{V. Piskovskyi et al.}

\institute{Università degli Studi di Milano-Bicocca, Milan, Italy \and
University of Essex, Colchester, England}

\maketitle

\begin{abstract}
The purpose of this work is to investigate the soundness and utility of a neural network-based approach as a framework for exploring the impact of image enhancement techniques on visual cortex activation. In a preliminary study, we prepare a set of state-of-the-art brain encoding models, selected among the top 10 methods that participated in The Algonauts Project 2023 Challenge \cite{gifford2023algonauts}. We analyze their ability to make valid predictions about the effects of various image enhancement techniques on neural responses. Given the impossibility of acquiring the actual data due to the high costs associated with brain imaging procedures, our investigation builds up on a series of experiments. Specifically, we analyze the ability of brain encoders to estimate the cerebral reaction to various augmentations by evaluating the response to augmentations targeting objects (i.e., faces and words) with known impact on specific areas. Moreover, we study the predicted activation in response to objects unseen during training, exploring the impact of semantically out-of-distribution stimuli. We provide relevant evidence for the generalization ability of the models forming the proposed framework, which appears to be promising for the identification of the optimal visual augmentation filter for a given task, model-driven design strategies as well as for AR and VR applications. 

\end{abstract}

\section{Introduction}
The historical connection between computer vision and brain modeling has long been a driving force in the field of artificial intelligence and machine learning. 
This interdisciplinary approach has been fueling advancements in both domains, drawing strong inspiration from our understanding of the brain's layered, distributed,  and  "active"  processing mechanisms \cite{marr2010vision, friston2010free, haxby2014decoding, yamins2016using, hassabis2017neuroscience}. 
Modern machine learning models have progressed beyond traditional computer vision tasks, now capable of predicting rich brain activation patterns based on presented natural stimuli \cite{naselaris2015resolving,kriegeskorte2019interpreting}. 
Although these models do not achieve spatially and temporally detailed predictions of brain responses, do not fully encapsulate the brain's active, embodied, and adaptive nature \cite{ognibene2014ecological,friston2015active,kessler2024human,tatler2011eye}, and lack substantial biological and anatomical constraints, they offer a valuable perspective for understanding brain function, in particular about learning and visual processing \cite{haxby2014decoding,kietzmann2017deep,richards2019deep,cichy2019deep}. These systems serve as \textit{data-driven brain models that maximize the likelihood of observed stimulus-response datasets}, thereby creating brain-inspired visual systems that replicate certain brain functions. 
This evolving synergy between computer vision and brain modeling underscores the potential of machine learning to enhance our understanding of both artificial and biological vision systems paving the way for advanced brain-computer interfaces \cite{bashivan2019neural,shenoy2021measurement}.  Furthermore, it opens up possibilities for identifying efficient and low-fatigue XR (Extended Reality) enhancement for specific tasks \cite{SALEM2019300} and developing neuroscience-informed design strategies, bridging the gap between computational models and practical applications in human-computer interaction \cite{gevins2003neurophysiological,martinez2020physiological,wickens2021engineering}.

In this context, image enhancement techniques play a crucial role in Computer-Aided Visual Perception, aiming to facilitate information extraction from visual stimuli. These techniques range from low-level approaches, such as color and brightness adjustments, to more complex methods that distinguish and highlight specific objects within images \cite{SALEM2019300}. While widely used in various fields, including medical imaging and augmented reality \cite{billinghurst2015survey}, understanding the impact of these enhancements on human visual processing is essential for improving their effectiveness. In particular, current computational principles ascribed to the brain seem to solve an inverse problem of reconstructing the sources of sensory input \cite{rao1999predictive,friston2010free, knill2004bayesian,vilares2011bayesian}. Surprisingly, a naive adoption of these \textit{computational principles would predict that computer aided visual perception would be more demanding} for the user having to reconstruct the role of a non ecological source, the computer. This calls for a more nuanced understanding of the underlying neural processes.
Traditional methods of studying these effects often involve expensive brain imaging techniques \cite{logothetis2008we,cinel2019neurotechnologies}. However, the advancements in machine learning and brain modeling discussed earlier offer a promising alternative for investigating the neural responses to enhanced visual stimuli, potentially reducing the need for expensive and complex procedures and opening new avenues for optimizing image enhancement strategies \cite{cichy2019deep}.

The aim of this work is to explore an alternative approach, which does not require conducting separate brain imaging procedures to study each of the numerous augmentation techniques selected for a specific task. This method consists in employing a set of neural networks with different characteristics, each pre-trained on a publicly available dataset that provides fMRI recordings of brain responses to non-augmented natural scene images. The key idea is that these brain encoding models can be used to predict the reaction of the visual cortex to novel stimuli that have been augmented with different techniques depending on the objective to be achieved. The advantage of not having to perform scanning procedures for each set of enhancement methods to test is evident. The main concern is, however, the reliability, soundness and performance of this approach. 

We initiate an exploration into these variables by analyzing the generalization ability of neural network models in predicting human brain responses to novel and enhanced visual stimuli. To achieve the highest possible accuracy in neural signal reconstruction from original, non-augmented images, we selected models from the top-ranked solutions in the renowned neuroscientific competition, The Algonauts Project 2023 \cite{gifford2023algonauts}. Due to the unavailability of actual brain responses to enhanced images, we estimate generalizability through a series of tests, pattern identification, and evaluation of agreement levels among different encoding models. Several results are also compared with basic principles from neuroscientific studies to validate our findings.

\section{Related  Work}

There are a variety of studies analyzing image enhancement techniques, typically with the aim of identifying the optimal technique, which explicitly rely on the human perception of these augmentations. Some of these approaches are focused solely on properties of the Human Visual System (HVS) known from theory, such as \cite{Chen:19}, while others involve more explicit models of visual mechanisms. For instance, \cite{Wharton, Wharton2, ZhangMultiScale} analyzed several enhancement algorithms, adopting a mathematical approach based on Logarithmic Image Processing \cite{Stockham,panetta} for simulating the HVS. However, explicit investigation of the impact of image enhancement on the human brain appears to be an under-explored area in this field. The problem of simulating how various image transformations are perceived by the visual cortex is largely unaddressed.

With the advent of Deep Neural Networks (DNN), their ability to learn both low-level and more complex visual features \cite{zeiler2013visualizingunderstandingconvolutionalnetworks}, proved to be useful for the brain encoding task. Numerous studies, conducted with the aim of investigating various brain processes, employed this method as a means to simulate and explain such mechanisms. For instance, \cite{EICKENBERG2017184} presented an original DNN-based fMRI encoder to investigate the characteristics of the Human Visual System by drawing parallels between its architecture and the hierarchical organization of Convolutional Neural Networks. \cite{Guclu10005} adopted a DNN to map natural images to neural signals with the objective of studying the representation of complex visual features in the human brain. An interesting conclusion reported in that study is that this approach, based on deep learning models, still has considerable drawbacks and is not capable of precisely replicating all the complex neural processes. Many other examples of studies, employing or investigating this category of brain encoding solutions, can be found. The specific models adopted in this work are
\cite{yang2023memory, Adeli2023.08.02.551743, nguyen2023algonauts, lane2023parameterefficient, matsuyama2023applicability, gifford2023algonauts}.

Creating a single framework for studying heterogeneous image enhancement techniques is challenging. Empirical studies have demonstrated that models with similar in-distribution performance can exhibit significantly different out-of-distribution performance \cite{McCoy, sun2019unsupervised}. This suggests that a model’s ability to generalize beyond its training data can vary greatly, and performance metrics within the training distribution may not accurately predict performance on novel data. However, foundation models, pretrained on large and diverse datasets, show promise in maintaining performance despite variations in image content (see \cite{bommasani2021opportunities,miller2021accuracy} for an in-depth discussion). Nevertheless, these models may not always mitigate distribution shifts, particularly those arising from spurious correlations \cite{sagawa2020investigation} or temporal changes \cite{radford2021learning, ke2021chextransfer}.

To investigate this issue, recent studies have employed datasets with induced shifts, such as corrupted images \cite{hendrycks2019benchmarking} or altered backgrounds \cite{xiao2020noise, sagawa2019distributionally}, to analyze model robustness. These methods provide valuable insights into how models handle variations and generalize to new conditions. By utilizing datasets with various augmentations, it is possible to assess how well brain encoding models generalize to different types of image transformations. This approach aligns with the broader goal of understanding and enhancing model robustness and generalization in the context of brain responses to visual stimuli.

\section{Methodology: Brain Encoding Models}

In order to develop the proposed framework upon high-performance brain encoding neural networks, we selected the best-ranked models that participated in The Algonauts Project 2023 Challenge \cite{gifford2023algonauts}. These solutions were built to accurately predict human brain responses to images of natural scenes, in the form of Blood Oxygenation Level Dependent (BOLD) functional Magnetic Resonance Imaging (fMRI) values observed in a series of voxels.

We followed the training procedures reported in the original papers of the $2^\mathrm{nd}$ \cite{Adeli2023.08.02.551743}, $3^\mathrm{rd}$ \cite{nguyen2023algonauts} and $6^\mathrm{th}$ \cite{lane2023parameterefficient} classified models. We also employed the baseline approach \cite{gifford2023algonauts} presented by the organizers. In order to increase the diversity of the set of neural networks, we included our original brain encoding model based on \textit{VMamba} \cite{liu2024vmamba} (Appendix). We plan also to test positional encoding based methods \cite{chimisso2023exploration}.
We trained the neural networks, using only the \textit{Subject 1} data from the Natural Scenes Dataset (NSD) \cite{allen2022nsd}, which was adopted in the Algonauts Challenge. Given that the training procedure of the $6^\mathrm{th}$ classified model does not provide for such a separation, we trained this network on the entire set of stimuli and extracted predictions using the module specific to \textit{Subject 1}. Due to hardware limitations and time constraints we were not able to exactly replicate the training recipes proposed by the authors. Specifically, in most cases, the ensembles of various configurations described in the original reports were not fully reproduced. Table 1 (Appendix) summarizes the trained models. The remaining methods from the top 10 were not considered due to the code unavailability. The winning \cite{yang2023memory} and the $8^\mathrm{th}$ classified were not used in later test due to the issues mentioned in Table 1 (Appendix).
We were able to assess the achievement of the reported performance for the $6^\mathrm{th}$ classified model only, mainly due to the fact that its authors reported the scores computed with a reproducible metric. 

We estimated the reliability of the models’ predictions by measuring their intrinsic entropy, which is independent of data diversity. As suggested in \cite{Gawlikowski2023} and previously shown in \cite{gustafsson2020evaluating}, the ensembling method exhibits superior performance as a model uncertainty metric. Its main point is to introduce as much variety as possible in the predictions made by each of the trained neural networks. Thus, we adopt the data randomization and shuffling approaches proposed by \cite{RENDA20191, Gawlikowski2023}, while a random initialization of weights is already implemented by default. We perform a 5-fold cross-validation, with each fold containing 1968 (20\%) randomly selected images from the training set of 9841 stimuli. We generate brain response predictions using 5 instances of each model and measure the mean voxel-wise standard deviation to estimate the entropy. As shown in Table 2 (Appendix), in addition to the highest correlation, the $6^\mathrm{th}$ classified model exhibits the lowest uncertainty level. We consider this model the most reliable.

\section{Results}
\subsection{Analysis of the predicted changes in the fMRI BOLD signal induced by image enhancement techniques}

\begin{figure}
\centering
\captionsetup[subfigure]{justification=centering}
  \begin{subfigure}[t]{0.16666\textwidth}
    \centering
    \captionsetup{justification=centering}
    \includegraphics[width=0.95\linewidth]{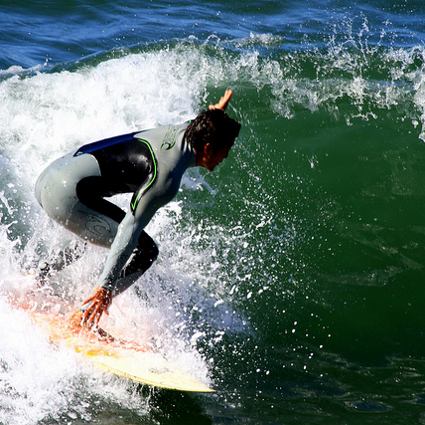}
    \caption{Original} \label{img:original}
  \end{subfigure}%
  \begin{subfigure}[t]{0.16666\textwidth}
    \centering
    \captionsetup{justification=centering}
    \includegraphics[width=0.95\linewidth]{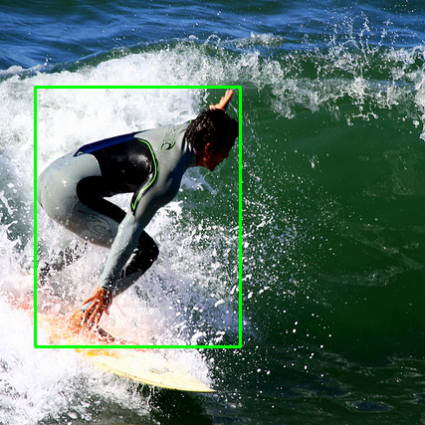}
    \caption{Bounding\\Box} \label{img:bb}
  \end{subfigure}%
  \begin{subfigure}[t]{0.16666\textwidth}
    \centering
    \captionsetup{justification=centering}
    \includegraphics[width=0.95\linewidth]{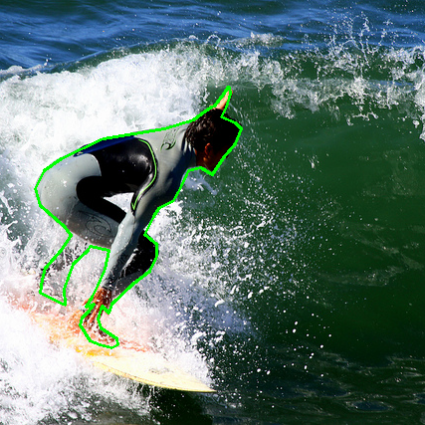}
    \caption{Contours} \label{img:contours}
  \end{subfigure}%
  \begin{subfigure}[t]{0.16666\textwidth}
    \centering
    \captionsetup{justification=centering}
    \includegraphics[width=0.95\linewidth]{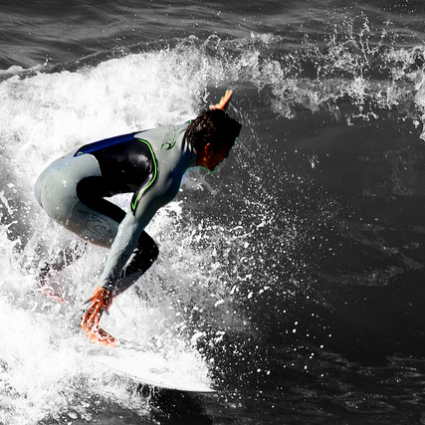}
    \caption{Grayscale} \label{img:gray}
  \end{subfigure}%
  \begin{subfigure}[t]{0.16666\textwidth}
    \centering
    \captionsetup{justification=centering}
    \includegraphics[width=0.95\linewidth]{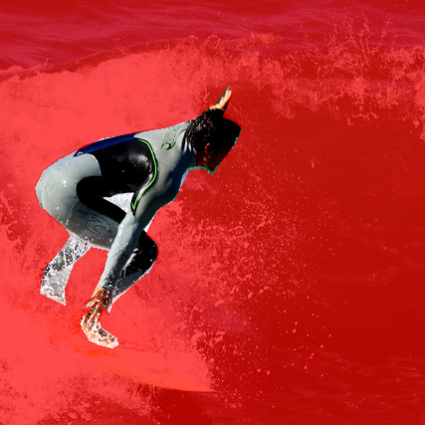}
    \caption{Inverse\\Overlay} \label{img:inv_overlay}
  \end{subfigure}%
  \begin{subfigure}[t]{0.16666\textwidth}
    \centering
    \captionsetup{justification=centering}
    \includegraphics[width=0.95\linewidth]{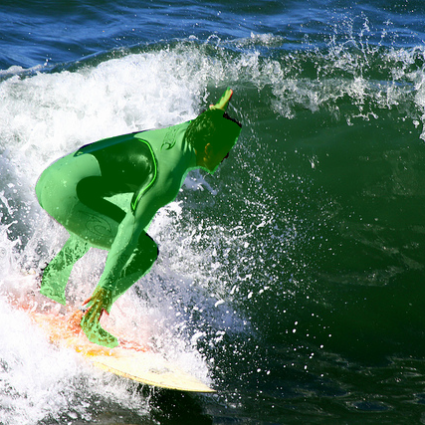}
    \caption{Overlay} \label{img:overlay}
  \end{subfigure}%
  \\
  \begin{subfigure}[t]{1\textwidth}
    \centering
    \captionsetup{justification=centering}
    \includegraphics[width=1\linewidth]{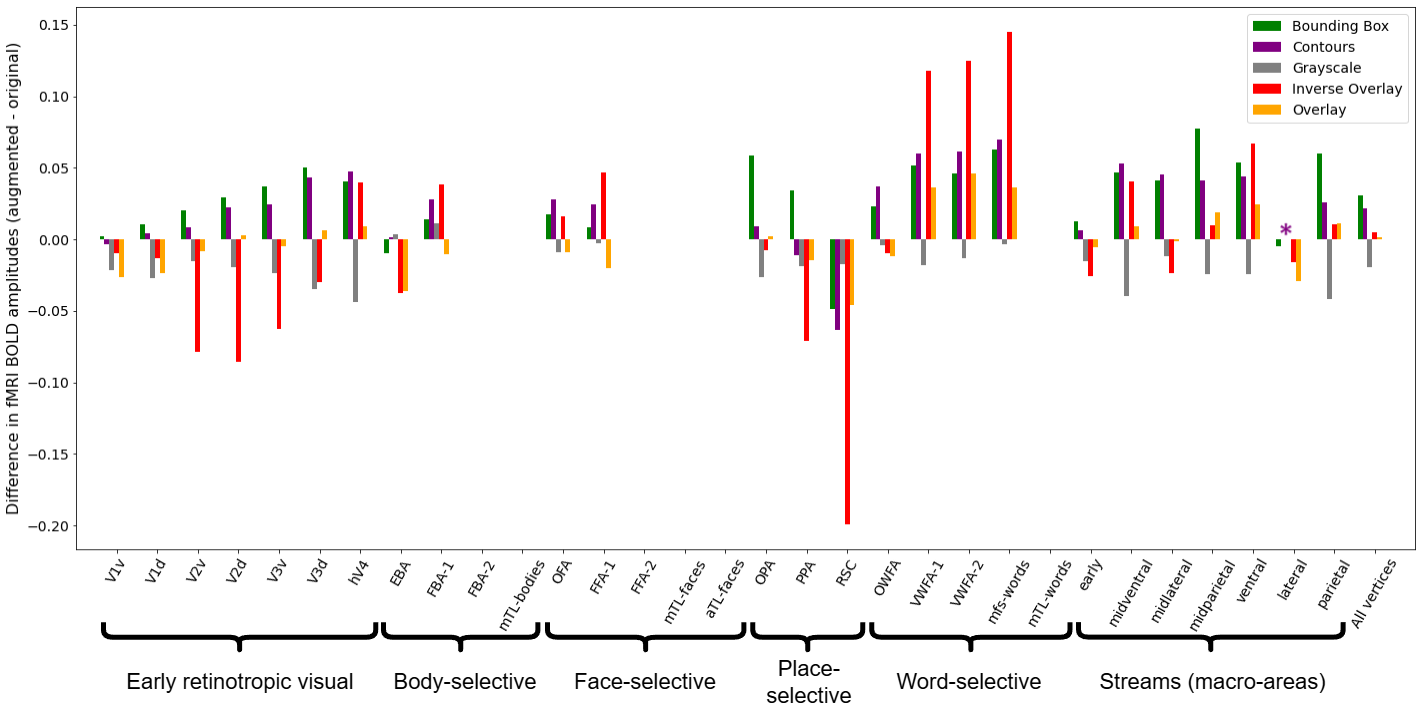}
    \caption{Predicted differences in activation} \label{img:differences_all_lh}
  \end{subfigure}%
  \hspace*{\fill} 

\caption{(a) - (f) Examples of different augmentations applied on the same image. \\(g) Differences between the neural responses predicted for the enhanced images and for their original versions. The values are averaged across 1956 images, 4 different models, and all voxels in each ROI of the left hemisphere only. (*) Refers to the only case in which an augmentation does not cause a statistically significant change in the signal according to the T-Test for the mean activation predicted for the original and augmented images.}
\label{img:differences_augmentations}
\end{figure}

We use all the 1956 images from the test sets of all the subjects from NSD to create 5 differently augmented versions of each of them (\Cref{img:original,img:bb,img:contours,img:gray,img:inv_overlay,img:overlay}). These augmentations aim at highlighting one or more objects, which correspond to the original annotated segments of the COCO dataset \cite{lin2015microsoftcococommonobjects} adopted by \cite{allen2022nsd}. We use all the trained models to predict visual cortex response to all the stimuli. Figure \ref{img:differences_all_lh} shows the predicted changes in the fMRI BOLD signal (the difference between the reaction to the enhanced stimulus and the response predicted for its original version), averaged across all the models, stimuli, and vertices in each Region of Interest (ROI).

\begin{figure}[!htb]
\centering
    \begin{subfigure}[t]{0.15\textwidth}
        \centering
        \captionsetup{justification=centering}
        \includegraphics[width=0.9\linewidth]{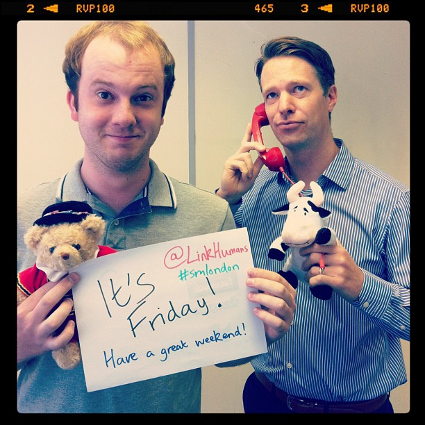}
        \caption{}
        \label{img:example_faces_a}
    \end{subfigure}%
    \begin{subfigure}[t]{0.15\textwidth}
        \centering
        \captionsetup{justification=centering}
        \includegraphics[width=0.9\linewidth]{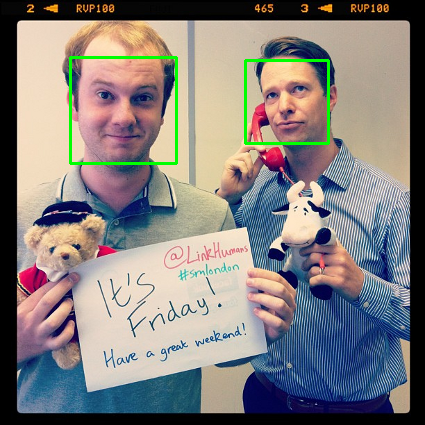}
        \caption{}
        \label{img:example_faces_b}
    \end{subfigure}%
    \begin{subfigure}[t]{0.15\textwidth}
        \centering
        \captionsetup{justification=centering}
        \includegraphics[width=0.9\linewidth]{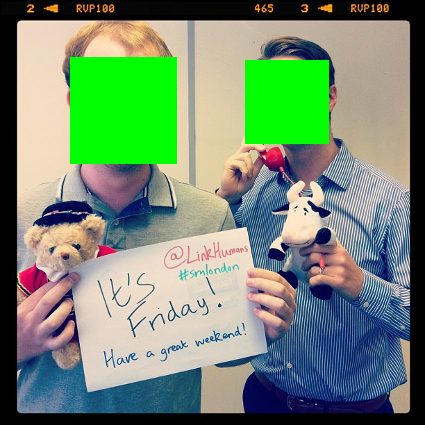}
        \caption{}
        \label{img:example_faces_c}
    \end{subfigure}%
    \begin{subfigure}[t]{0.15\textwidth}
        \centering
        \captionsetup{justification=centering}
        \includegraphics[width=0.9\linewidth]{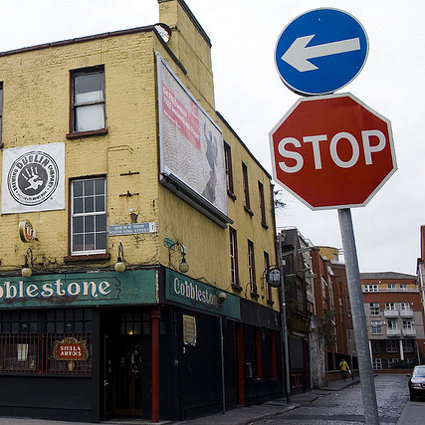}
        \caption{}
        \label{img:example_words_a}
    \end{subfigure}%
    \begin{subfigure}[t]{0.15\textwidth}
        \centering
        \captionsetup{justification=centering}
        \includegraphics[width=0.9\linewidth]{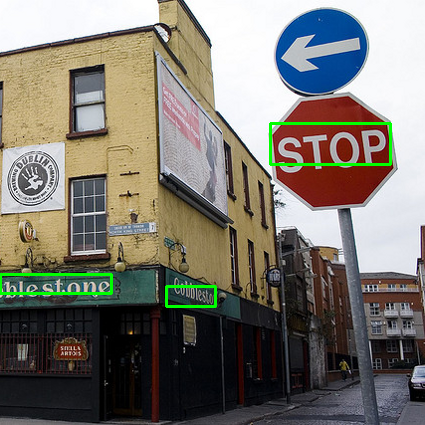}
        \caption{}
        \label{img:example_words_b}
    \end{subfigure}%
    \begin{subfigure}[t]{0.15\textwidth}
        \centering
        \captionsetup{justification=centering}
        \includegraphics[width=0.9\linewidth]{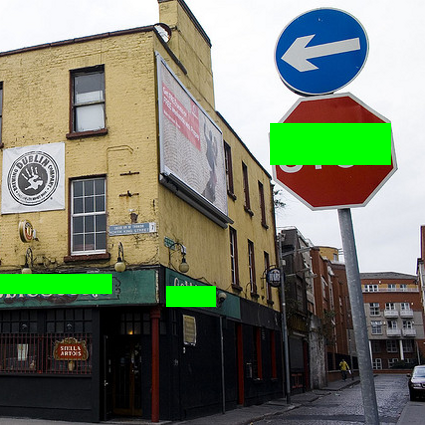}
        \caption{}
        \label{img:example_words_c}
    \end{subfigure}%


\caption[Examples of images with differently augmented human faces and words.]{Examples of images with differently augmented human faces (a-c) and words (d-f). a,d: original; b,e: highlighted with bounding box;  c,f: covered. } 
\label{img:example_faces_and_words}
\end{figure}

To have a better understanding of the reliability of the predictions in this out of distribution setting, we analyzed the model’s performance on particular classes of augmentations: text and faces 
(chosen due to the presence of the corresponding brain areas in the dataset \cite{allen2022nsd}), as shown in Figure \ref{img:example_faces_and_words}. We observed that the models perform reasonably well on these augmentations. In fact, as illustrated in the plots in Figures 13-16 in the Appendix, the brain regions associated with these specific types of object are sensitive to their occlusion (i.e., when they are obscured). This sensitivity highlights the ability of the model to accurately predict neural responses to altered stimuli. Since the neural correlates of faces and text are well-established, the model’s effective performance on these objects suggests its overall robustness and reliability in predicting brain responses to various augmentations.

\begin{table}[]
\footnotesize
\centering

\resizebox{\linewidth}{!}{
\begin{tabular}{|c| c c| c c| c c? c c| c c| c c|| c c| c c|} 
 \hline
 \multirow[b]{3}{*}{\thead{\makecell{\textbf{Enhancement}\\\textbf{Technique}}}} & 
 \multicolumn{6}{c?}{\thead{\makecell{\textbf{Signal}}}} &
 \multicolumn{10}{c|}{\thead{\makecell{\textbf{Signal Variation (prediction augm. - prediction orig.)}}}} \\
 \cline{2-17}
 & \multicolumn{2}{c|}{\thead{\makecell{\textbf{SD}}}} & 
   \multicolumn{2}{c|}{\thead{\makecell{\textbf{SD across}\\\textbf{all images}}}} & 
   \multicolumn{2}{c?}{\thead{\makecell{\textbf{SD across}\\\textbf{all folds}}}} & 
   \multicolumn{2}{c|}{\thead{\makecell{\textbf{SD}}}} & 
   \multicolumn{2}{c|}{\thead{\makecell{\textbf{SD across}\\\textbf{all images}}}} & 
   \multicolumn{2}{c||}{\thead{\makecell{\textbf{SD across}\\\textbf{all folds}}}} & 
   \multicolumn{2}{c|}{\thead{\makecell{\textbf{Mean abs.}\\\textbf{diff}}}} & 
   \multicolumn{2}{c|}{\thead{\makecell{\textbf{Max abs.}\\\textbf{diff}}}} \\
 & LH & RH & LH & RH & LH & RH & LH & RH & LH & RH & LH & RH & LH & RH & LH & RH
 
 \\
 \hline

    \textit{Original (none)} & 0.279 & 0.279 & 0.279 & 0.279 & 0.050 & 0.050 & 0 & 0 & 0 & 0  & 0 & 0 & 0 & 0 &0 &0\\
    \hline    
    \textit{Bounding Box} & 
    $0.257$ & $0.256$ & $0.257$ & $0.256$ & $\textbf{0.048}$ & $\textbf{0.048}$ & $0.047$ & $0.048$ & $0.045$ & $0.047$ & $\textbf{0.013}$ & $\textbf{0.013}$ & $0.043$ & $0.044$ & $0.142$ & $0.146$ \\

    \textit{Contours} & 
    $0.263$ & $0.262$ & $0.263$ & $0.262$ & $\textbf{0.049}$ & $\textbf{0.049}$ & $0.037$ & $0.038$ & $0.037$ & $0.038$ & $\textbf{0.010}$ & $\textbf{0.010}$ & $0.035$ & $0.036$ & $0.122$ & $0.123$ \\
    
    \textit{Grayscale} & 
    $0.267$ & $0.268$ & $0.267$ & $0.267$ & $\textbf{0.050}$ & $\textbf{0.050}$ & $0.052$ & $0.053$ & $0.050$ & $0.050$ & $\textbf{0.020}$ & $\textbf{0.019}$ & $0.051$ & $0.053$ & $0.232$ & $0.282$ \\
    
    \textit{Inverse Overlay} & 
    $0.243$ & $0.245$ & $0.241$ & $0.242$ & $\textbf{0.052}$ & $\textbf{0.052}$ & $0.106$ & $0.104$ & $0.100$ & $0.099$ & $\textbf{0.040}$ & $\textbf{0.040}$ & $0.119$ & $0.116$ & $0.442$ & $0.417$ \\
    
    \textit{Overlay} & 
    $0.258$ & $0.256$ & $0.257$ & $0.256$ & $\textbf{0.048}$ & $\textbf{0.048}$ & $0.048$ & $0.049$ & $0.047$ & $0.048$ & $\textbf{0.012}$ & $\textbf{0.012}$ & $0.040$ & $0.041$ & $0.135$ & $0.148$ \\
\hline
\end{tabular}}
\caption{Columns 2-13: Mean voxel-wise standard deviation of both the fMRI BOLD signal predicted by the $6^\mathrm{th}$ classified model and its variation. The standard deviation is computed across predictions for 1956 testing images made by 5 different instances of this network, each trained on a separate fold, comprising 20\% of all the available training images (for \textit{Subject 1}). All the values are averaged across all the vertices in the corresponding hemisphere. Columns 14-17: Absolute differences between the responses predicted for the same image with and without augmentations.}
\label{tab:std_6th_folds}
\end{table}

To further evaluate the robustness of the models' predictions we trained the $6^\mathrm{th}$ classified model on 5 different folds, comprising 20\% of all the images for \textit{Subject 1}. We observe that the mean standard deviation across all the images is higher than the same metric computed over all the images (for a single model or fold). This means that the variation of the predicted responses is mainly due to the intrinsic variability of the stimuli, indicating a high consistency of the predictions made by this network. We also notice that although the standard deviation is low, the model remains sensible to augmentations, exhibiting a significant variation of predicted activation (columns 14-17 of Table \ref{tab:std_6th_folds}). In the appendix we report the area specific analysis.

\subsection{Models trained without images of specific object classes}

First, considering a relatively high performance of both the $3^\mathrm{rd}$ and $6^\mathrm{th}$ classified models in the previous experiment, we perform the training of these models on the original dataset, from which we separately remove images depicting cats, dogs, mobile phones, and TVs (two natural entities and two man-made objects). Second, we train the same brain encoders, excluding only 100 stimuli for each category, in order to use them as the test set. In the latter case we also eliminate some randomly selected pictures to obtain two datasets of the same size for each type of object. We evaluate the error of the predictions made by the two versions of models on the test set containing 100 images, each depicting the corresponding entities (cats, dogs, mobile phones, TVs). For each of the obtained instances (8 in total) of the original dataset, the $6^\mathrm{th}$ classified model is also trained on 5 randomly generated folds, each containing 20\% of all images (remaining after the mentioned removal of certain stimuli). We evaluate the consistency of the predictions by computing the mean voxel-wise standard deviation across the 5 instances, each trained on a separate fold. These results are summarized in Table \ref{tab:training_without_objects}. In the appendix we report the area specific analysis.   

\begin{table}[!ht]
\scriptsize
\centering
\begin{tabular}{|c| c c| c c| c c| c c| c| c|} 
 \hline
 \multirow[c]{2}{*}{\thead{\makecell{\textbf{Configuration}}}} & 
 \multicolumn{2}{c|}{\thead{\makecell{\textbf{MAE}}}} & 
 \multicolumn{2}{c|}{\thead{\makecell{\textbf{SD}}}}  & 
 \multicolumn{2}{c|}{\thead{\makecell{\textbf{SD across}\\\textbf{100 images}}}}  & 
 \multicolumn{2}{c|}{\thead{\makecell{\textbf{SD across}\\\textbf{5 folds}}}}  & 
 \multirow[c]{2}{*}{\thead{\makecell{\textbf{Corr.}\\\textbf{Subj01}}}} &
 \multirow[c]{2}{*}{\thead{\makecell{\textbf{Total}\\\textbf{images}}}}  \\ 
 & LH & RH & LH & RH & LH & RH & LH & RH & & 
 \\
 \specialrule{2.5pt}{1pt}{1pt}
\multicolumn{11}{|c|}{\makecell{$\textbf{3}^\mathrm{\textbf{rd}}$ \textbf{classified model}}} \\ \hline
 
    \textit{Phones not seen} & 
    $\textbf{0.486}$ & $\textbf{0.489}$ & $0.349$ & $0.347$ & $0.348$ & $0.346$ & $0.107$ & $0.109$ & 
    $\textbf{0.442}$ & \multirow[c]{2}{*}{355} \\
    \textit{Phones seen} & 
    $0.489$ & $0.492$ & $\textbf{0.342}$ & $\textbf{0.339}$ & $\textbf{0.341}$ & $\textbf{0.338}$ & $\textbf{0.099}$ & $\textbf{0.100}$ & $0.431$ &\\ \hline

    \textit{TVs not seen} & 
    $\textbf{0.488}$ & $\textbf{0.491}$ & $0.300$ & $0.301$ & $0.298$ & $0.300$ & $0.107$ & $0.109$ & $\textbf{0.441}$ &
    \multirow[c]{2}{*}{317} \\
    \textit{TVs seen} & 
    $0.491$ & $0.496$ & $\textbf{0.291}$ & $\textbf{0.293}$ & $\textbf{0.290}$ & $\textbf{0.292}$ & $\textbf{0.091}$ & $\textbf{0.095}$ & $0.433$ & \\ \hline

    \textit{Cats not seen} & 
    $0.499$ & $0.507$ & $0.265$ & $0.266$ & $0.263$ & $0.264$ & $0.101$ & $0.104$ & $\textbf{0.434}$ & 
    \multirow[c]{2}{*}{363} \\
    \textit{Cats seen} & 
    $\textbf{0.497}$ & $\textbf{0.506}$ & $\textbf{0.261}$ & $\textbf{0.262}$ & $\textbf{0.260}$ & $\textbf{0.260}$ & $\textbf{0.099}$ & $\textbf{0.101}$ & $0.434$ &\\ \hline

    \textit{Dogs not seen} & 
    $\textbf{0.514}$ & $\textbf{0.519}$ & $0.318$ & $0.324$ & $0.317$ & $0.322$ & $0.103$ & $0.107$ & $\textbf{0.434}$ & 
    \multirow[c]{2}{*}{374} \\
    \textit{Dogs seen} & 
    $0.515$ & $0.519$ & $\textbf{0.315}$ & $\textbf{0.319}$ & $\textbf{0.315}$ & $\textbf{0.318}$ & $\textbf{0.099}$ & $\textbf{0.102}$ & $0.432$ &\\ \hline

\multicolumn{11}{|c|}{\makecell{$\textbf{6}^\mathrm{\textbf{th}}$ \textbf{classified model}}} \\ \hline
 
    \textit{Phones not seen} & 
    $0.468$ & $0.467$ & $\textbf{0.220}$ & $\textbf{0.216}$ & $\textbf{0.219}$ & $\textbf{0.215}$ & $0.040$ & $0.041$ & 
    $0.500$ & \multirow[c]{2}{*}{355} \\
    \textit{Phones seen} & 
    $\textbf{0.466}$ & $\textbf{0.466}$ & $0.223$ & $0.219$ & $0.222$ & $0.218$ & $\textbf{0.040}$ & $\textbf{0.039}$ & $\textbf{0.505}$ &\\ \hline

    \textit{TVs not seen} & 
    $0.470$ & $0.469$ & $\textbf{0.173}$ & $\textbf{0.174}$ & $\textbf{0.172}$ & $\textbf{0.172}$ & $\textbf{0.039}$ & $0.039$ & $0.503$ &
    \multirow[c]{2}{*}{317} \\
    \textit{TVs seen} & 
    $\textbf{0.469}$ & $\textbf{0.469}$ & $0.177$ & $0.177$ & $0.176$ & $0.176$ & $0.039$ & $\textbf{0.039}$ & $\textbf{0.506}$ & \\ \hline

    \textit{Cats not seen} & 
    $0.484$ & $0.486$ & $\textbf{0.157}$ & $\textbf{0.154}$ & $\textbf{0.155}$ & $\textbf{0.152}$ & $\textbf{0.039}$ & $\textbf{0.039}$ & $0.504$ & 
    \multirow[c]{2}{*}{363} \\
    \textit{Cats seen} & 
    $\textbf{0.482}$ & $\textbf{0.485}$ & $0.160$ & $0.156$ & $0.158$ & $0.154$ & $0.043$ & $0.042$ & $\textbf{0.505}$ &\\ \hline

    \textit{Dogs not seen} & 
    $\textbf{0.491}$ & $\textbf{0.492}$ & $\textbf{0.191}$ & $\textbf{0.191}$ & $\textbf{0.190}$ & $\textbf{0.191}$ & $\textbf{0.039}$ & $\textbf{0.039}$ & $0.503$ & 
    \multirow[c]{2}{*}{374} \\
    \textit{Dogs seen} & 
    $0.492$ & $0.494$ & $0.196$ & $0.198$ & $0.195$ & $0.197$ & $0.040$ & $0.040$ & $\textbf{0.504}$ &\\ \hline
\end{tabular}
\small
\caption{Comparison of the models trained with and without images of various objects/animals. Column 1: Configuration "not seen" refers to the removal of all the images of a specific object from the training set. Configuration "seen" refers to the removal of only 100 of such pictures. Columns 2: Mean Voxel-wise Absolute Error. Columns 3-5: Standard Deviation across 5 instances of the model (each trained on 20\% of the images in the corresponding dataset) and/or across all 100 test images. Column 6: Mean voxel-wise Pearson's correlation achieved. Column 7: Total number of images depicting the corresponding objects in the original dataset.}
\label{tab:training_without_objects}
\end{table}

\section{Conclusion}

The generalization ability of the analyzed neural networks and, thus, the validity of the proposed framework as a means of studying image enhancement techniques are supported by relevant evidence. Some sensible and explainable patterns are observed when evaluating and comparing various changes in the fMRI signal predicted by the models as a response to augmentations applied to visual stimuli. 
Specifically, the impact of some augmentations is found to be consistent with the results of neuroscientific studies concerning object selectivity (the reactions to individual images observed in certain brain areas) and color sensitivity of brain regions (the responses of early visual ROIs to transformations affecting colors). Relatively low variability of predictions indicates low uncertainty of the models regarding the expected impact of augmentations. For most of the areas of the visual cortex, several neural networks (and in some cases all of them) agree on the changes induced by various enhancement techniques, which suggests the soundness of the proposed framework.

Furthermore, the generalizability of the fMRI encoders is supported by the results of a series of experiments aimed at verifying the sensitivity of certain brain regions to specific objects depicted in images. In ROIs known to be word-selective and face-selective, sensible activation patterns were revealed when evaluating their response to augmentations applied on text and human faces, respectively. The predicted partial overlap between the functions of the two brain regions and differences between the reactions observed in the left and right hemispheres appear to be consistent with some neuroscientific studies \cite{Hagen2020-sh, Behrmann2013-jh}. Additional experiments are conducted to investigate the role of the saliency of image areas, dependence of prediction variability on the number of altered pixels, and several other aspects. Their results, reported in the Appendix, suggest the validity of the obtained neural responses. A further contribution to the assessment of generalizability is provided by the similarly high performance exhibited by the encoders when predicting the responses to \textit{out-of-distribution} objects, which were not seen during training. 

The achieved results confirm that the proposed approach can be useful for estimating the effects of image enhancement algorithms on the human perception of visual stimuli. This framework is also expected to be useful in AR and VR applications where the utilization of certain filters may generate particularly useful patterns of neural activation. This may help improve the performance of human agents in certain activities, such as learning to perform a new task.

%
%
\bibliographystyle{splncs04}
\bibliography{main}

\end{document}